\documentclass[11pt,a4paper]{article}
\usepackage[hyperref]{emnlp2020}
\usepackage{times}
\usepackage{latexsym}

\usepackage{microtype}

\aclfinalcopy 
\def\aclpaperid{1383} 

\usepackage{amsfonts}
\usepackage{bm}
\usepackage{array,amsmath}
\usepackage{amssymb}
\usepackage{dsfont}
\usepackage{float}
\usepackage{graphicx}
\usepackage{algorithm}
\usepackage{algorithmicx}
\usepackage{algpseudocode}
\usepackage{pgf,tikz}
\usepackage{mathrsfs}
\usepackage{multirow}
\usepackage{booktabs}
\usetikzlibrary{arrows}

\def\figref#1{Figure~\ref{fig:#1}}
\def\figlabel#1{\label{fig:#1}\label{p:#1}}

\def\tabref#1{Table~\ref{tab:#1}}
\def\tablabel#1{\label{tab:#1}\label{p:#1}}

\def\secref#1{\S\ref{sec:#1}}
\def\seclabel#1{\label{sec:#1}}
\def\eqref#1{Eq.~\ref{eqn:#1}}

\newcounter{notecounter}
\newcommand{\enotesoff}{\long\gdef\enote##1##2{}}
\newcommand{\enoteson}{\long\gdef\enote##1##2{{
			\stepcounter{notecounter}
			{\large\bf
				\hspace{1cm}\arabic{notecounter} $<<<$ ##1: ##2
				$>>>$\hspace{1cm}}}}}
\enoteson
\enotesoff
\long\def\eat#1{}

\title{Identifying Elements Essential for BERT's Multilinguality}

\author{Philipp Dufter, Hinrich Sch\"{u}tze\\
	Center for Information and Language Processing (CIS), LMU Munich, Germany\\
	{\tt philipp@cis.lmu.de}}

\date{}

\begin{document}
\maketitle

\begin{abstract}
 It has been shown that multilingual BERT (mBERT) yields high quality multilingual representations 
and enables effective zero-shot transfer. 
This is surprising given that mBERT does not use any  crosslingual signal during training.
While  recent literature has studied this phenomenon,
the reasons for the multilinguality are still somewhat obscure.
We aim to identify architectural properties of BERT
and linguistic properties of languages that are necessary
for BERT to become multilingual.
To allow for fast experimentation we propose an efficient
setup with small BERT models trained on a mix of synthetic and
natural  data.
Overall, we identify four architectural and two linguistic elements
that influence multilinguality.
Based on our insights,
we experiment with
a multilingual pretraining setup
that  modifies the masking strategy using VecMap, i.e., unsupervised embedding alignment.
Experiments on XNLI with three languages indicate that our findings transfer from our small setup
to larger scale settings.

\end{abstract}

\section{Introduction}

Multilingual models, i.e., models capable of processing
more than one language  with comparable performance, 
are central to natural language processing. They are useful as fewer models
need to be maintained to serve many languages, resource requirements are reduced, 
and low- and mid-resource languages can benefit from crosslingual transfer.
Further, multilingual models are useful in machine translation, zero-shot task transfer and 
typological research.
There is a clear need 
for 
multilingual models
for the world's 7000+ languages.

\begin{figure}[t]
	\centering
	\includegraphics[width=1.0\linewidth]{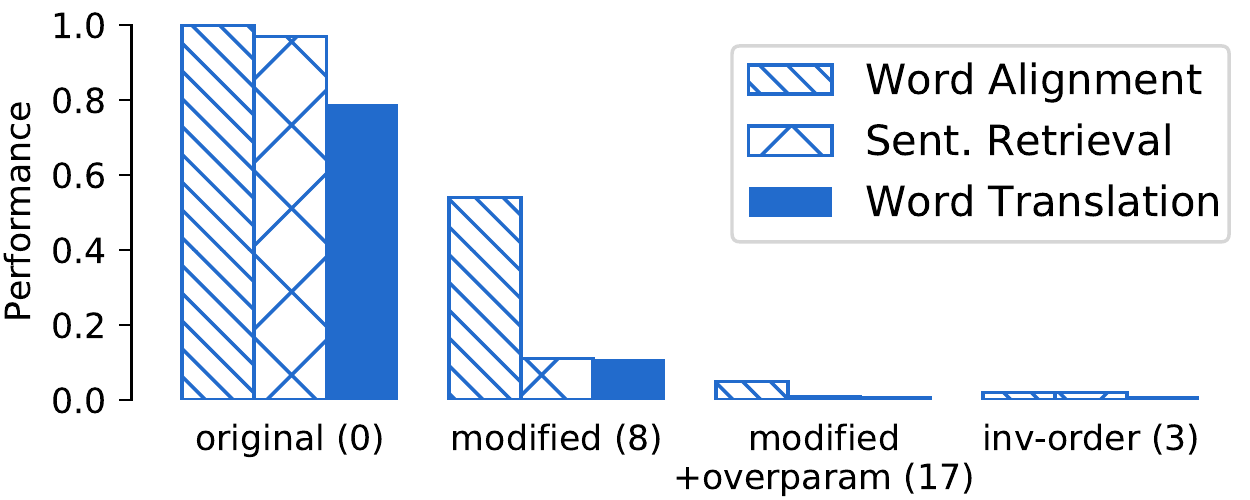}
	\caption{Multilinguality in our BERT model (0) is harmed by
          three architectural modifications: lang-pos,
          shift-special, no-random (8); see
          \secref{archprop} for definitions.
           Together with overparameterization
		almost no multilinguality is left (17).
Pairing a language with its inversion (i.e., inverted word order)
destroys multilinguality as well (3). Having parallel training corpora is helpful for
 multilinguality (not shown).
Results are for embeddings from layer 8.}
	\figlabel{summary}
\end{figure}

With the rise of static word embeddings,
many multilingual embedding algorithms have been proposed
\cite{mikolov2013exploiting,hermann-blunsom-2014-multilingual,faruqui-dyer-2014-improving}; for a survey see \cite{ruder2019survey}.
Pretrained language models
\cite{peters-etal-2018-deep,howard-ruder-2018-universal,devlin-etal-2019-bert}
have high performance across 
tasks, outperforming static word embeddings. 
A simple multilingual model is multilingual BERT\footnote{\url{https://github.com/google-research/bert/blob/master/multilingual.md}} (\emph{mBERT}).
It is a BERT-Base model \cite{devlin-etal-2019-bert} trained on the 104 largest Wikipedias with a shared subword vocabulary. There is no additional crosslingual signal. 
Still, mBERT yields high-quality multilingual representations \cite{pires-etal-2019-multilingual,wu2019beto,hu2020xtreme}.

The exact reason for mBERT's multilinguality is -- to the best of our knowledge -- still debated. 
\citet{wang2019cross} provide an extensive study and conclude that a shared vocabulary
is not necessary, but that the model needs to be deep and languages need to share a 
similar ``structure''. \citet{artetxe2019cross} show that neither a shared vocabulary nor
joint pretraining is required for BERT to be multilingual. \citet{wu2019emerging} 
find that BERT models across languages can be easily aligned
and that a necessary requirement for achieving multilinguality are shared parameters in the top layers. This work continues this line of research. We find indications
that six elements influence the multilinguality of BERT.
\figref{summary} summarizes  our main findings.

\subsection{Contributions}
\begin{itemize}
	\item  Training BERT models consumes tremendous resources. We propose an experimental
	 setup that allows for fast experimentation. 
	 \item We hypothesize that BERT is multilingual because of a limited number of parameters. By forcing 
	 the model to use its parameters efficiently, it exploits common structures by aligning 
	 representations across languages.  We provide experimental 
	 evidence that the number of parameters and training duration is interlinked with multilinguality
	 and an indication that generalization and multilinguality might be conflicting goals.
	\item   We show that shared special tokens, shared position embeddings and the common masking strategy
	to replace masked tokens with random words contribute to multilinguality. This is in line with findings from \cite{wu2019emerging}.
	\item   We show that having identical structure across languages, but an inverted word order in one language
	destroys multilinguality. Similarly having shared position embeddings contributes to multilinguality. We thus hypothesize that word order across languages is an important ingredient for multilingual models.
	\item   Using these insights we perform initial experiments to create a model with higher degree of multilinguality.
	\item 
        We conduct experiments on Wikipedia and evaluate
	on XNLI to show that our findings transfer to larger scale settings.
\end{itemize}

Our code is publicly available.\footnote{\url{https://github.com/pdufter/minimult}}

\section{Setup and Hypotheses}

\subsection{Setup}
We aim at 
having a setup that allows for gaining insights quickly when 
investigating multilinguality. Our assumption is that these insights are transferable 
 to a larger scale real world setup. We verify this assumption in \secref{real}.

\textbf{Languages.}
\citet{wang2019cross} propose to consider English and Fake-English, 
a language that is created by shifting unicode points by a large constant. 
Fake-English in their case has the exact same linguistic properties as English, but is 
represented by different unicode points.  We follow a similar approach, but instead of shifting unicode points we simply
shift token indices after tokenization by a constant;
shifted tokens are prefixed by ``::'' and added to the vocabulary.
See \figref{fakeeng} for an example.
While shifting indices and unicode code points have similar effects, we
chose shifting indices as we find it somewhat cleaner.\footnote{For example, the BERT tokenizer treats 
some punctuation as special symbols  (e.g., ``dry-cleaning'' is tokenized as
[``dry'', ``-'', ``\#\#cleaning''], not as [``dry'', ``\#\#-'', ``\#\#cleaning'']). 
When using a unicode shift,
tokenizations of English and Fake-English can differ.}

\begin{figure}[t]
	\centering
	\includegraphics[width=1.0\linewidth]{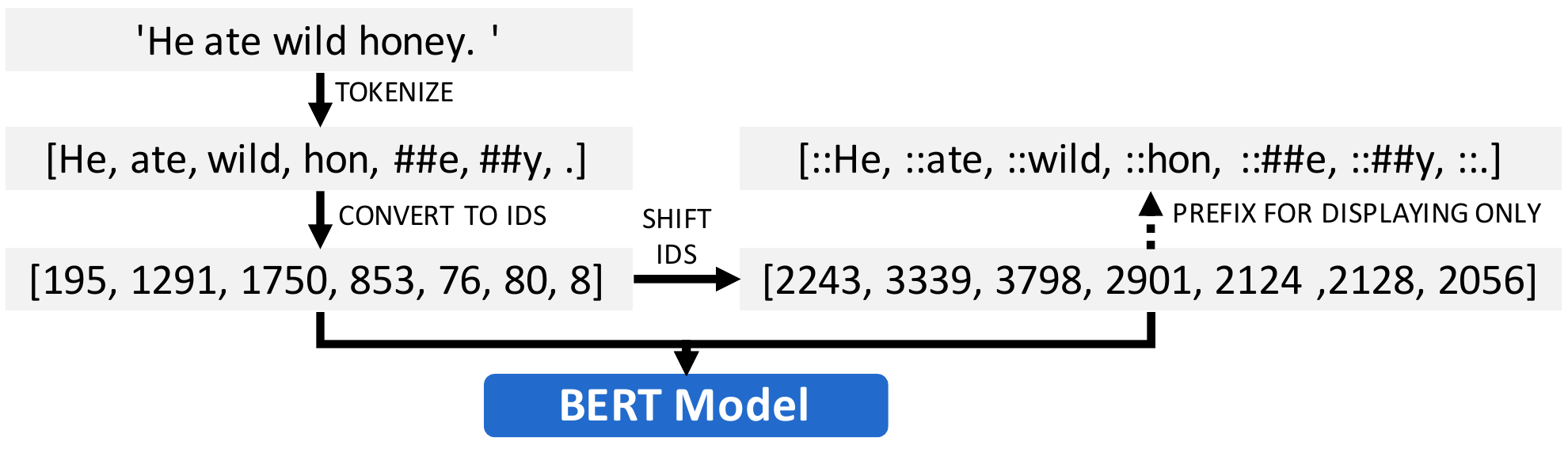}
	\caption{Creating a Fake-English sentence by adding a shift of 2048 to token indices.}
	\figlabel{fakeeng}
\end{figure}

\textbf{Data.}
For our setup, aimed at supporting fast experimentation,
a small corpus with limited vocabulary is desirable.
As training data we
use the English Easy-to-Read version of the Parallel Bible Corpus \cite{mayer2014creating}
 that contains the New Testament.
 The corpus is structured into verses and is word-tokenized.
We sentence-split verses using NLTK \cite{loper-bird-2002-nltk}. 
The final corpus has 17k sentences, 228k words, a vocabulary size of 4449 and 71 distinct characters. 
The median sentence length is 12 words. By creating a Fake-English version of this corpus we get a shifted replica and thus 
a sentence-parallel corpus.

As development data we apply the same procedure to the first 10k sentences of the Old Testament of the English
King James Bible. All our evaluations are performed on development data, except for word translation and 
when indicated explicitly.

\textbf{Vocabulary.}
We create a vocabulary of size 2048 from the Easy-to-Read Bible with the wordpiece tokenizer \cite{schuster2012japanese}.\footnote{\url{https://github.com/huggingface/tokenizers}}
Using the same vocabulary for English and Fake-English yields
a final vocabulary size of 4096.

\textbf{Model.}
We use the BERT-Base architecture
\cite{devlin-etal-2019-bert}, modified to achieve a smaller model: we divide hidden size,
  intermediate size of the feed forward layer and number of
  attention heads by 12; thus,
 hidden size is 64 and intermediate size 256. While this leaves us with a single attention head, 
\citet{wang2019cross} found that the number 
 of attention heads is important neither for  overall performance nor for  multilinguality. We call this smaller model \emph{BERT-small}.

 As a consistency check for our experiments 
 we consider random embeddings in the form of a randomly initialized but untrained
 BERT model, referred to as \textbf{\emph{``untrained''}}.

 \textbf{Training Parameters.}
 We mostly use the original training parameters as given in \cite{devlin-etal-2019-bert}. 
 Learning rate and number of epochs was chosen to achieve reasonable perplexity on the training corpus 
 (see supplementary for details). 
Unless indicated differently we use a batch size of 256, train for 100 epochs with AdamW \cite{loshchilov2018fixing} 
(learning rate 2e-3, weight decay .01, epsilon 1e-6), and use 50 warmup steps. 
We only use the masked-language-modeling objective, without
next-sequence-prediction.
With this setup we can train a single model in under 40 minutes on a single 
GPU (GeForce GTX 1080Ti). We run each experiment with five different seeds, and report 
mean and standard deviation. 

\subsection{Evaluation}

We evaluate two properties of our trained language models: the \emph{degree of multilinguality}
and -- as a consistency check -- the \emph{overall model fit} (i.e., is the trained language 
model 
of reasonable quality). 

\subsubsection{Multilinguality}
We evaluate the degree of multilinguality with three tasks.
Representations from different layers of BERT can be considered. 
We use layer 0 (uncontextualized) and layer 8 
(contextualized). Several papers have found layer 8 to work well for monolingual
and multilingual tasks \cite{tenney-etal-2019-bert,hewitt-manning-2019-structural,sabet2020simalign}.
Note that representations from layer 0 include position and segment embeddings besides the token embeddings as well as 
layer normalization.

\textbf{Word Alignment.}
\citet{sabet2020simalign} find that mBERT performs well on
word alignment.
By construction, we have a sentence-aligned corpus with English 
and Fake-English. The gold word alignment between two sentences is the 
identity alignment. 
We use this automatically 
created gold-alignment for evaluation.

To extract word alignments from BERT we use \cite{sabet2020simalign}'s Argmax method.
Consider the parallel sentences $s^{(\text{eng})},s^{(\text{fake})}$,
with length $n$. We extract $d$-dimensional wordpiece embeddings from the 
$l$-th layer of BERT to obtain embeddings $\mathcal{E}(s^{(k)}) \in \mathbb{R}^{n \times d}$
for $k \in \{\text{eng}, \text{fake}\}$. The similarity matrix $S \in [0,1]^{n \times n}$ is 
computed by $S_{ij} := \text{cosine-sim}\left(\mathcal{E}(s^{(\text{eng})})_i, \mathcal{E}(s^{(\text{fake})})_j\right)$.
Two wordpieces $i$ and $j$ are aligned if
\begin{align*}
(i = \arg\max_l S_{l,j}) \wedge  (j = \arg\max_l S_{i,l}).
\end{align*}

The alignments are evaluated using precision, recall and $F_1$ as follows: 
\begin{equation*}
\text{p} = \frac{|P \cap G|}{|P|}, \;
\text{r} = \frac{|P \cap G|}{|G|}, \;
F_1 = \frac{2 \; \text{p}\; \text{r}}{\text{p} + \text{r}},
\end{equation*}
where $P$ is the set of predicted alignments and $G$ the set of true alignment edges.
We report $F_1$.

\textbf{Sentence Retrieval}
is popular for evaluating crosslingual representations (e.g., \cite{artetxe2019massively,libovicky2019language}).
We obtain the embeddings $\mathcal{E}(s^{(k)})$ as
before and compute
a sentence embedding $e_s^{(k)}$ simply by averaging vectors across all tokens in a 
sentence (ignoring CLS and SEP tokens). Computing cosine 
similarities between English and Fake-English sentences
yields the similarity matrix $R \in \mathbb{R}^{m \times m}$ 
where $R_{ij} = \text{cosine-sim}(e_i^{(\text{eng})}, e_j^{(\text{fake})}) $
for $m$ sentences.

Given an English query sentence $s_i^{(\text{eng})}$, we obtain the retrieved sentences in Fake-English by 
ranking them according to similarity. Since we can do
the same with Fake-English
as query language, we report the mean precision of these directions, computed as
$$
\rho = \frac{1}{2m}\sum_{i = 1}^{m}\mathds{1}_{\arg\max_{l}R_{il} = i} + \mathds{1}_{\arg\max_{l}R_{li} = i}.
$$

We also evaluate
\textbf{word translation.}
Again, by construction we have a ground-truth bilingual dictionary of size 2048.
We obtain word vectors by feeding each word in the vocabulary
individually to BERT, in the form
``[CLS] \{token\} [SEP]''.
We then evaluate word translation like sentence retrieval and denote the measure 
with $\tau$.

\textbf{Multilinguality Score.}
For an easier overview we compute a multilinguality score by averaging
retrieval and translation results across both layers. That is 
$\mu = 1 / 4(\tau_0 + \tau_8 + \rho_0 + \rho_8)$ where
$\tau_k$,$\rho_k$ means representations from layer $k$ have been used.
We omit word
alignment here
as it is not a suitable measure to compare  all models: with
shared position embeddings,
the task is almost trivial given that the gold alignment is the identity alignment.

\subsubsection{Model Fit}
\textbf{MLM Perplexity.}
To verify that BERT was successfully trained we evaluate the models
on perplexity (with base $e$) for training and development data.
Perplexity is computed on 15\% of randomly selected tokens that are replaced
by ``[MASK]''.
Given those randomly selected tokens in a text ${w_1, \dots, w_n}$ and
probabilities ${p_{w_1}, \dots, p_{w_n}}$ that the correct token was predicted by the model, 
perplexity is calculated as $\exp(-1/n \sum_{k = 1}^{n} \log(p_{w_k})).$

\subsection{Architectural Properties}
\seclabel{archprop}

Here we formulate hypotheses as to which architectural components
contribute to multilinguality.

\textbf{Overparameterization: \emph{overparam}.}
If BERT is severely overparameterized
the model should 
have enough capacity to model each language separately without creating a multilingual space.
Conversely, if the number of parameters is small, the model 
has a need to use parameters efficiently. The model is likely to identify
common structures among  languages and model them together, 
thus creating a multilingual space.

To test this, we train a larger BERT model
that has the same configuration as BERT-base  (i.e., hidden size: 768, intermediate size: 3072, attention heads: 12) 
and is thus much larger than our standard configuration, BERT-small. 
Given our small training corpus and the small number of languages,
we argue that BERT-base is overparameterized.
For the overparameterized model we use learning rate 1e-4 (following \cite{devlin-etal-2019-bert}).

\textbf{Shared Special Tokens: \emph{shift-special}.}
It has been found that a shared vocabulary is not essential for multilinguality 
\cite{wang2019cross,artetxe2019cross,wu2019emerging}.
Similar to prior studies, in our setting each language has its own vocabulary, as we aim at breaking the multilinguality of BERT. However in prior studies, special tokens ([UNK], [CLS], [SEP], [MASK],
[PAD]) are usually shared 
across languages. 
Shared special tokens may contribute to multilinguality
because
they are very frequent and could serve as
``anchor points''. To investigate this,
we shift the special tokens with the same shift as applied to token indices.

\textbf{Shared Position Embeddings: \emph{lang-pos}.}
Position and segment embeddings are usually shared across languages. We investigate their contribution
to multilinguality by using language-specific position (\emph{lang-pos}) and segment embeddings.
For an example see \figref{lspecpos}.

\begin{figure}[t]
	\centering
	\includegraphics[width=1.0\linewidth]{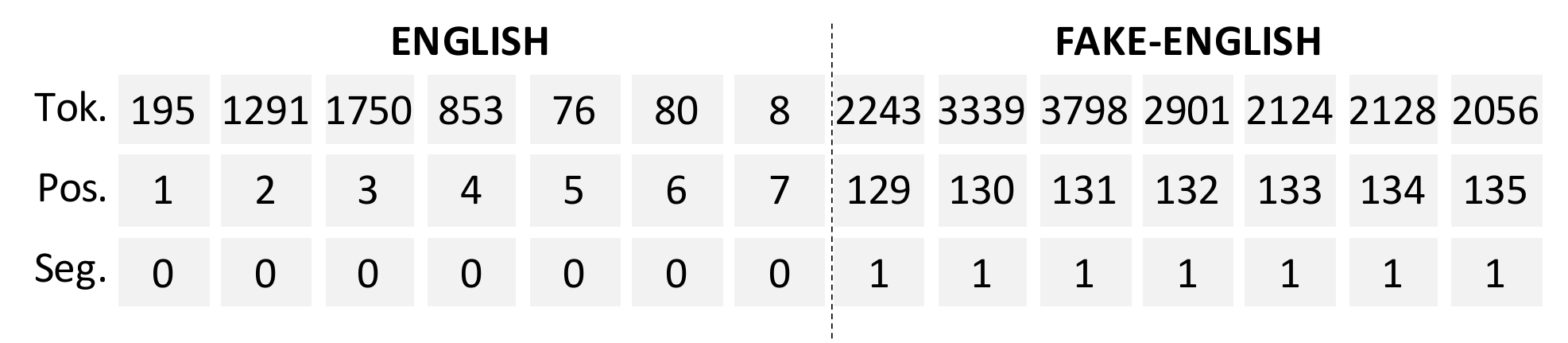}
	\caption{lang-pos: input indices to BERT with language specific position and segment embeddings.}
	\figlabel{lspecpos}
\end{figure}

\textbf{Random Word Replacement: \emph{no-random}.}
The MLM task as proposed by \citet{devlin-etal-2019-bert} masks 15\% 
of tokens in a sentence.
These tokens are replaced with ``[MASK]''
in $p_{\text{[mask]}}=80\%$,
remain unchanged in $p_{\text{[id]}}=10\%$ and are
replaced with a random token of the vocabulary in $p_{\text{[rand]}}=10\%$ of the cases.
The randomly sampled token can come from any language resulting in Fake-English tokens 
 to appear in English sentences and vice-versa. 
We hypothesize that this random replacement could contribute
to multilinguality.
We experiment with
the setting
$p=(0.8, 0.2, 0.0)$ where
$p$ denotes the triple
$(p_{\text{[mask]}},
p_{\text{[id]}}, p_{\text{[rand]}})$.

\subsection{Linguistic Properties}

\textbf{Inverted Word Order: \emph{inv-order}.}
\citet{wang2019cross} shuffled word order in sentences randomly and found that word order has some, but not a severe effect on multilinguality. 
They conclude that ``structural similarity'' across languages is important without further specifying this term.
We investigate an extreme case: inversion.
We invert
each sentence in the Fake-English corpus: $[w_1, w_2,
  \ldots, w_n] \rightarrow [w_n, w_{n-1}, \ldots, w_1]$.
Note that,
apart from the reading order, all properties of the languages are preserved, including 
ngram statistics. Thus, the structural similarity of
English and inverted Fake-English is
arguably
very high.

\textbf{Comparability of Corpora: \emph{no-parallel}.}
We hypothesize that
the similarity
of  training corpora contributes to ``structural similarity'':  if we train on a parallel corpus we expect the language structures
to be more similar than when we train on two independent corpora, potentially from different domains. 
For mBERT,  Wikipedias across languages are  in the same domain, share some articles and thus
are comparable, yet not parallel. To test our hypothesis, we train on a non-parallel corpus.
We create it by splitting the Bible into two halves, using one half for English and Fake-English each, 
thus avoiding any parallel sentences during training.

\section{Results}

\subsection{Architectural Properties}
\begin{figure}[t]
	\centering
	\includegraphics[width=0.7\linewidth]{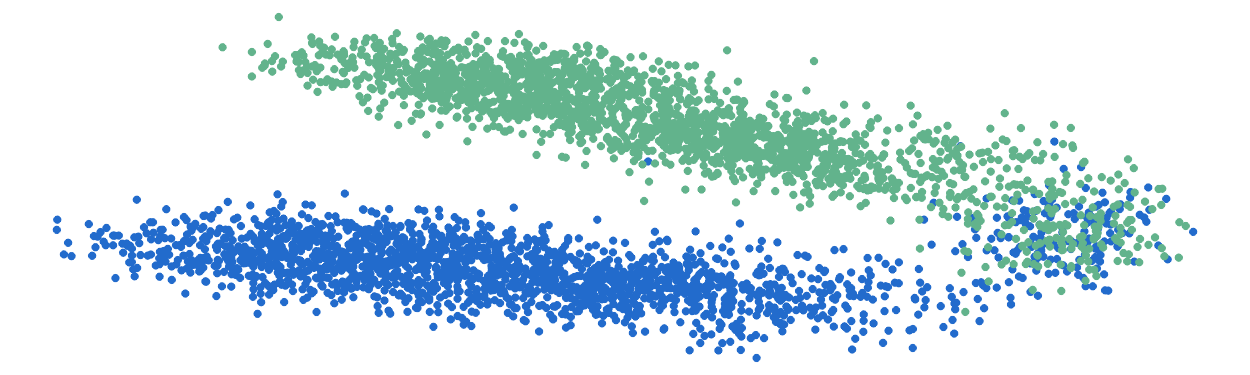}
	\includegraphics[width=0.9\linewidth]{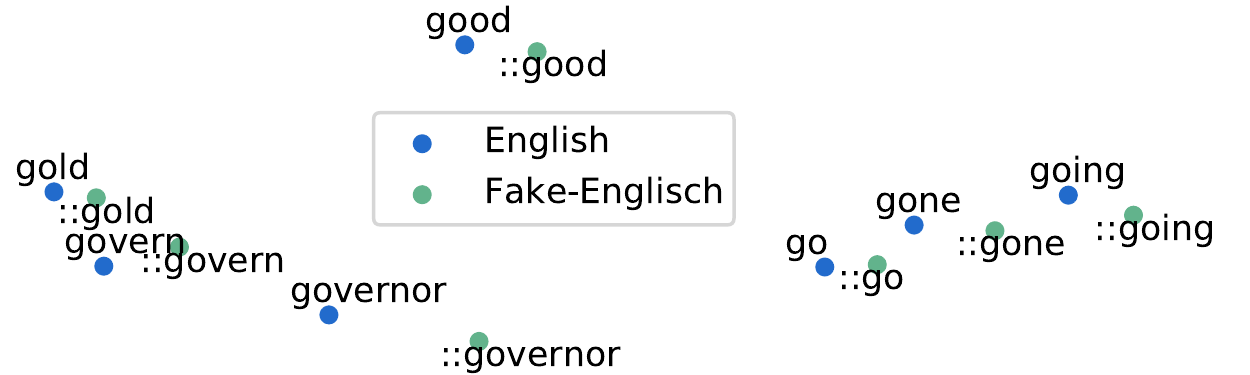}
	\caption{\emph{Top}:
          PCA of
          the \textbf{token embeddings} from layer 0
	  of the original model (ID 0).
          The representations
	of the two languages clearly have a similar
        structure. \emph{Bottom}:
        PCA of a sample of \textbf{token embeddings}.
        Corresponding tokens in English and Fake-English are
        nearest neighbors of each other or nearly so.
This is quantitatively confirmed in \tabref{main}.}
	\figlabel{pca}
\end{figure}

\begin{figure}[t]
	\centering
	\includegraphics[width=0.8\linewidth]{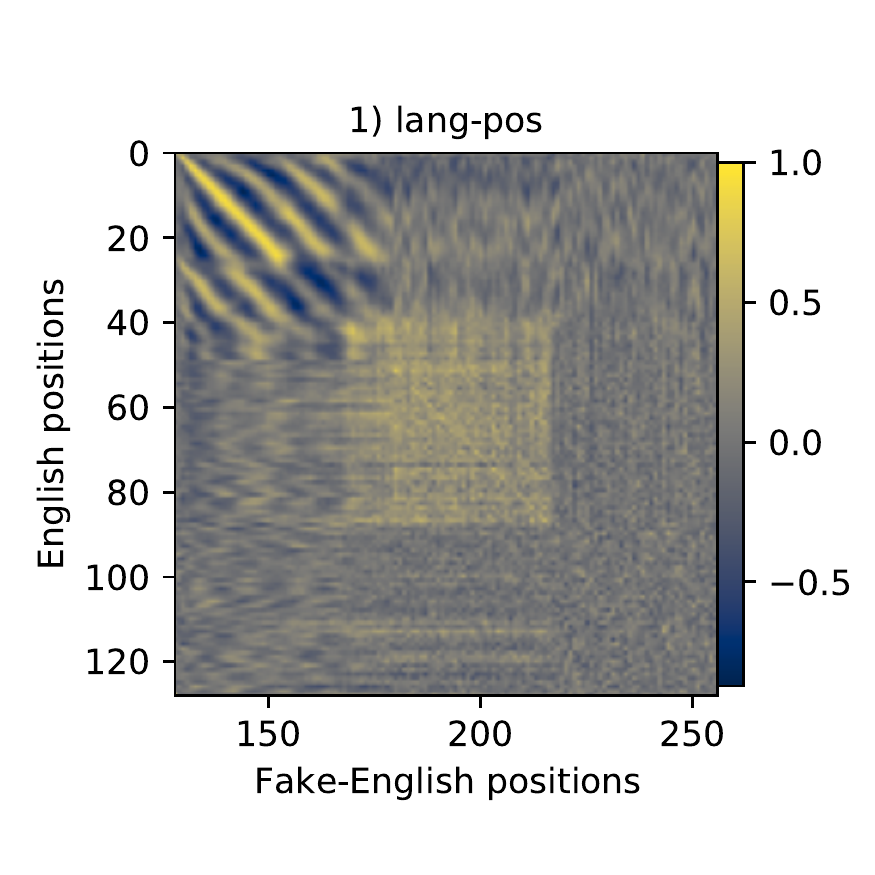}
	\includegraphics[width=0.8\linewidth]{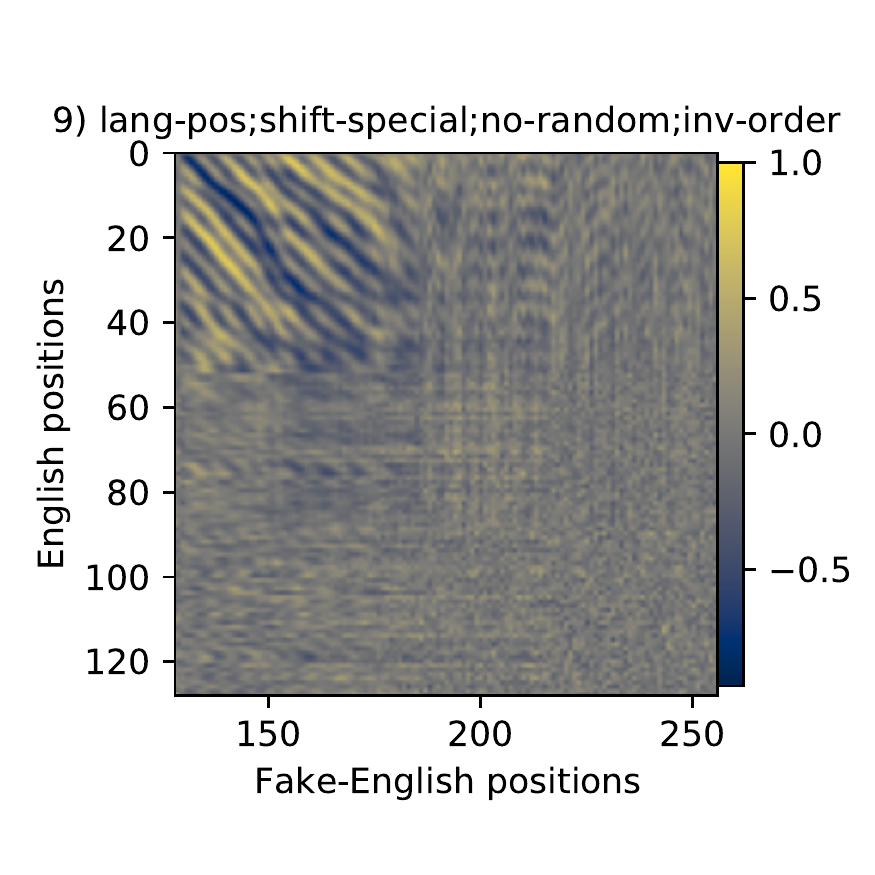}
	\caption{Cosine similarity matrices of position embeddings. The maximum length after tokenization
		in our experiments is 128. Position embedding IDs 0-127 are used
		by English, 128-255 by Fake-English.}
	\figlabel{posfig}
\end{figure}

\begin{table*}[t]
	\centering
	\scriptsize
	\def\symmsep{0.2cm}
	\begin{tabular}{@{\hspace{\symmsep}}r@{\hspace{\symmsep}}l||r|rrr|rrr||rr}
		&&  \textbf{Mult.-} & \multicolumn{3}{c}{ \textbf{Layer 0} } & \multicolumn{3}{c}{ \textbf{Layer 8} }  & \multicolumn{2}{c}{ \textbf{MLM-} } \\
		& & \textbf{score} & \textbf{Align.} & \textbf{Retr.} & \textbf{Trans.} & \textbf{Align.} & \textbf{Retr.} & \textbf{Trans.} &\multicolumn{2}{c}{ \textbf{Perpl.} }  \\
		\textbf{ID}&\textbf{Description} & \textbf{$\mu$} & \textbf{$F_1$} & \textbf{$\rho$} & \textbf{$\tau$} &  \textbf{$F_1$} & \textbf{$\rho$} & \textbf{$\tau$} &\textbf{train} & \textbf{dev}\\
		\midrule
		\midrule
		0 & original & .70 & 1.00 \textsubscript{.00} & .16 \textsubscript{.02} & .88 \textsubscript{.02} & 1.00 \textsubscript{.00} & .97 \textsubscript{.01} & .79 \textsubscript{.03} & 9 \textsubscript{\textcolor{white}{0}0.2} & 217 \textsubscript{\textcolor{white}{0}7.8} \\
		\midrule 1 & lang-pos & .30 & .87 \textsubscript{.05} & .33 \textsubscript{.13} & .40 \textsubscript{.09} & .89 \textsubscript{.05} & .39 \textsubscript{.15} & .09 \textsubscript{.05} & 9 \textsubscript{\textcolor{white}{0}0.1} & 216 \textsubscript{\textcolor{white}{0}9.0} \\
		2 & shift-special & .66 & 1.00 \textsubscript{.00} & .15 \textsubscript{.02} & .88 \textsubscript{.01} & 1.00 \textsubscript{.00} & .97 \textsubscript{.02} & .63 \textsubscript{.13} & 9 \textsubscript{\textcolor{white}{0}0.1} & 227 \textsubscript{17.9} \\
		4 & no-random & .68 & 1.00 \textsubscript{.00} & .19 \textsubscript{.03} & .87 \textsubscript{.02} & 1.00 \textsubscript{.00} & .85 \textsubscript{.07} & .82 \textsubscript{.04} & 9 \textsubscript{\textcolor{white}{0}0.6} & 273 \textsubscript{\textcolor{white}{0}7.7} \\
		5 & lang-pos;shift-special & .20 & .62 \textsubscript{.19} & .22 \textsubscript{.19} & .27 \textsubscript{.20} & .72 \textsubscript{.22} & .27 \textsubscript{.21} & .05 \textsubscript{.04} & 10 \textsubscript{\textcolor{white}{0}0.5} & 205 \textsubscript{\textcolor{white}{0}7.6} \\
		6 & lang-pos;no-random & .30 & .91 \textsubscript{.04} & .29 \textsubscript{.10} & .36 \textsubscript{.12} & .89 \textsubscript{.05} & .32 \textsubscript{.15} & .25 \textsubscript{.12} & 10 \textsubscript{\textcolor{white}{0}0.4} & 271 \textsubscript{\textcolor{white}{0}8.6} \\
		7 & shift-special;no-random & .68 & 1.00 \textsubscript{.00} & .21 \textsubscript{.03} & .85 \textsubscript{.01} & 1.00 \textsubscript{.00} & .89 \textsubscript{.06} & .79 \textsubscript{.04} & 8 \textsubscript{\textcolor{white}{0}0.3} & 259 \textsubscript{15.6} \\
		8 & lang-pos;shift-special;no-random & .12 & .46 \textsubscript{.26} & .09 \textsubscript{.09} & .18 \textsubscript{.22} & .54 \textsubscript{.31} & .11 \textsubscript{.11} & .11 \textsubscript{.13} & 10 \textsubscript{\textcolor{white}{0}0.6} & 254 \textsubscript{15.9} \\
		\midrule 15 & overparam & .58 & 1.00 \textsubscript{.00} & .27 \textsubscript{.03} & .63 \textsubscript{.05} & 1.00 \textsubscript{.00} & .97 \textsubscript{.01} & .47 \textsubscript{.06} & 2 \textsubscript{\textcolor{white}{0}0.1} & 261 \textsubscript{\textcolor{white}{0}4.5} \\
		16 & lang-pos;overparam & .01 & .25 \textsubscript{.10} & .01 \textsubscript{.00} & .01 \textsubscript{.00} & .37 \textsubscript{.13} & .01 \textsubscript{.00} & .00 \textsubscript{.00} & 3 \textsubscript{\textcolor{white}{0}0.0} & 254 \textsubscript{\textcolor{white}{0}4.9} \\
		17 & lang-pos;shift-special;no-random;overparam & .00 & .05 \textsubscript{.02} & .00 \textsubscript{.00} & .00 \textsubscript{.00} & .05 \textsubscript{.04} & .00 \textsubscript{.00} & .00 \textsubscript{.00} & 1 \textsubscript{\textcolor{white}{0}0.0} & 307 \textsubscript{\textcolor{white}{0}7.7} \\
		\midrule 3 & inv-order & .01 & .02 \textsubscript{.00} & .00 \textsubscript{.00} & .01 \textsubscript{.00} & .02 \textsubscript{.00} & .01 \textsubscript{.01} & .00 \textsubscript{.00} & 11 \textsubscript{\textcolor{white}{0}0.3} & 209 \textsubscript{14.4} \\
		9 & lang-pos;inv-order;shift-special;no-random & .00 & .04 \textsubscript{.01} & .00 \textsubscript{.00} & .00 \textsubscript{.00} & .03 \textsubscript{.01} & .00 \textsubscript{.00} & .00 \textsubscript{.00} & 10 \textsubscript{\textcolor{white}{0}0.4} & 270 \textsubscript{20.1} \\
		\midrule \midrule 18 & untrained & .00 & .97 \textsubscript{.01} & .00 \textsubscript{.00} & .00 \textsubscript{.00} & .96 \textsubscript{.01} & .00 \textsubscript{.00} & .00 \textsubscript{.00} & 3484 \textsubscript{44.1} & 4128 \textsubscript{42.7} \\
		19 & untrained;lang-pos & .00 & .02 \textsubscript{.00} & .00 \textsubscript{.00} & .00 \textsubscript{.00} & .02 \textsubscript{.00} & .00 \textsubscript{.00} & .00 \textsubscript{.00} & 3488 \textsubscript{41.4} & 4133 \textsubscript{50.3} \\
		\midrule \midrule 30 & knn-replace & .74 & 1.00 \textsubscript{.00} & .31 \textsubscript{.08} & .88 \textsubscript{.00} & 1.00 \textsubscript{.00} & .97 \textsubscript{.01} & .81 \textsubscript{.01} & 11 \textsubscript{\textcolor{white}{0}0.3} & 225 \textsubscript{12.4} \\
	\end{tabular}
	\caption{Multilinguality and model fit for our models. Mean and standard deviation (subscript)
		across 5 different random seeds is shown. ID is a unique identifier for the model setting.
		To put perplexities into perspective: the pretrained mBERT has a perplexity of roughly 46
		on train and dev. knn-replace is explained in \secref{improving}. 
		\tablabel{main}}
\end{table*}

\begin{table}[t]
	\centering
	\scriptsize
	\def\symmsep{0.08cm}
	\begin{tabular}{
			@{\hspace{\symmsep}}r@{\hspace{\symmsep}}
			@{\hspace{\symmsep}}l@{\hspace{\symmsep}}||
			@{\hspace{\symmsep}}r@{\hspace{\symmsep}}|
			@{\hspace{\symmsep}}r@{\hspace{\symmsep}}
			@{\hspace{\symmsep}}r@{\hspace{\symmsep}}
			@{\hspace{\symmsep}}r@{\hspace{\symmsep}}|
			@{\hspace{\symmsep}}r@{\hspace{\symmsep}}
			@{\hspace{\symmsep}}r@{\hspace{\symmsep}}
			@{\hspace{\symmsep}}r@{\hspace{\symmsep}}|
			@{\hspace{\symmsep}}r@{\hspace{\symmsep}}
			@{\hspace{\symmsep}}r@{\hspace{\symmsep}}}
		&&  & \multicolumn{3}{c}{ \textbf{Layer 0} } & \multicolumn{3}{c}{ \textbf{Layer 8} }  & \multicolumn{2}{c}{ \textbf{Perpl.} } \\
		\textbf{ID}&\textbf{Description} & \textbf{$\mu$} & \textbf{$F_1$} & \textbf{$\rho$} & \textbf{$\tau$} &  \textbf{$F_1$} & \textbf{$\rho$} & \textbf{$\tau$} &\textbf{train} & \textbf{dev}\\
		\midrule
		0 & original & .70 & 1.00  & .16  & .88  & 1.00  & .97  & .79  & 9  & 217  \\
		21 & no-parallel & .25 & .98  & .06  & .28  & .98  & .50  & .15  & 14  & 383  \\
		21b & lang-pos;no-parallel & .07 & .60  & .10  & .07  & .73  & .11  & .02  & 16  & 456  \\
	\end{tabular}
	\caption{Results showing the effect of having a parallel vs. non-parallel training corpus. \tablabel{parallel}}
\end{table}

\tabref{main} shows results. Each model has an associated ID 
that is consistent with the code.
The original model (ID 0) shows a high degree 
of multilinguality. As mentioned, alignment is an easy task with shared position embeddings yielding $F_1=1.00$. Retrieval works better with contextualized
representations on layer 8 ($.97$ vs.\ $.16$) whereas word
translation works better on layer 0 ($.88$ vs.\ $.79$), as expected. Overall the embeddings
seem to capture the similarity of English and Fake-English exceptionally well (see \figref{pca}
for a PCA of token embeddings).
The untrained BERT models perform poorly (IDs 18, 19), except for word alignment with shared position embeddings.

When applying our \textbf{architectural modifications}
(lang-pos, shift-special, no-random) individually
we see medium to slight decreases in multilinguality (IDs 1, 2, 4). lang-pos has the largest negative impact. 
Apparently,
applying just a single modification can be  compensated by the model. Indeed, 
when using two modifications at a time (5--7) multilinguality
goes down more, only with 7 there is still a high degree of multilinguality. With all three modifications (8) the degree of multilinguality is drastically lowered ($\mu$ $.12$ vs.\ $.70$).

We see
that the language model 
quality
(see columns MLM-Perpl.)
is stable on train and dev across  models (IDs 1--8) and does not
deviate
from original BERT (ID 0) by much.\footnote{Perplexities on dev are high because the
  English of the King
  James Bible
  is quite different from that of the Easy-to-Read
  Bible. Our research question is: which modifications harm
  BERT's multilinguality without harming model fit (i.e.,
  perplexity). The relative change of perplexities, not their
  absolute value is important in this context.}
Thus, we can conclude that each of the models has fitted the
training data well and poor results on $\mu$ are not due to the fact that
the architectural changes have  hobbled BERT's
language modeling performance.

 The \textbf{overparameterized} model (ID 15) exhibits lower scores for word translation, but higher ones for retrieval and
overall a lower multilinguality score (.58 vs. .70). 
However, when we add lang-pos (16) or apply all three architectural modifications
(17), multilinguality drops  to $.01$ and $.00$.
This indicates that 
by decoupling languages with the proposed modifications
(lang-pos, shift-special, no-random)
and greatly increasing the number of parameters (overparam), it is
possible to get a
well-performing language model
(low perplexity)
that is not multilingual. \emph{Conversely, we can conclude
that
the four architectural properties together are  necessary
for  BERT
to be multilingual.}

\subsection{Linguistic Properties}
Inverting Fake-English (IDs 3, 9)
 breaks multilinguality almost completely -- independently of any
architectural modifications. Having a language
with the exact same structure (same ngram statistics,
vocabulary size etc.), only with 
inverted order, seems to block BERT from creating a
multilingual space. Note that perplexity is almost the same.
\emph{We conclude that having a similar word order structure is necessary for BERT 
to create a multilingual space.} The fact that shared position embeddings are important for multilinguality supports this finding.
 Our hypothesis is that the drop in multilinguality with inverted word order
comes from an incompatibility between word and position encodings: BERT needs to learn that the 
word at position 0 in English is similar to word at position $n$ in Fake-English. 
However, $n$ (the sentence length) varies from sentence to
sentence.
This suggests that
 relative position embeddings -- rather than absolute
 position embeddings -- might be beneficial for multilinguality
 across languages with high distortion.

To investigate this effect more, \figref{posfig} shows cosine similarities between position embeddings for models 1, 9.
Position IDs 0-127 are for English, 128-255  for Fake-English. Despite
language specific position embeddings, the embeddings 
exhibit a similar structure: in the top panel
there is a clear yellow diagonal at the beginning, which weakens at the end.
The
bottom shows 
that for a model with inverted Fake-English the position
embeddings live in different spaces: no diagonal is visible.

In the range 90--128
(a rare sentence length)
the similarities look random.
This indicates that smaller position embeddings
are trained more than larger ones (which occur less frequently).
We suspect
that embedding similarity correlates with the number of gradient updates a single position embedding receives. 
Positions 0, 1 and 128, 129 receive a gradient update in every step and can thus be considered an average of all
gradient updates (up to random initialization).
This is potentially one reason for the diagonal 
pattern in the top panel.

\subsection{Corpus Comparability}

\begin{figure*}[t]
	\centering
	\includegraphics[width=0.45\linewidth]{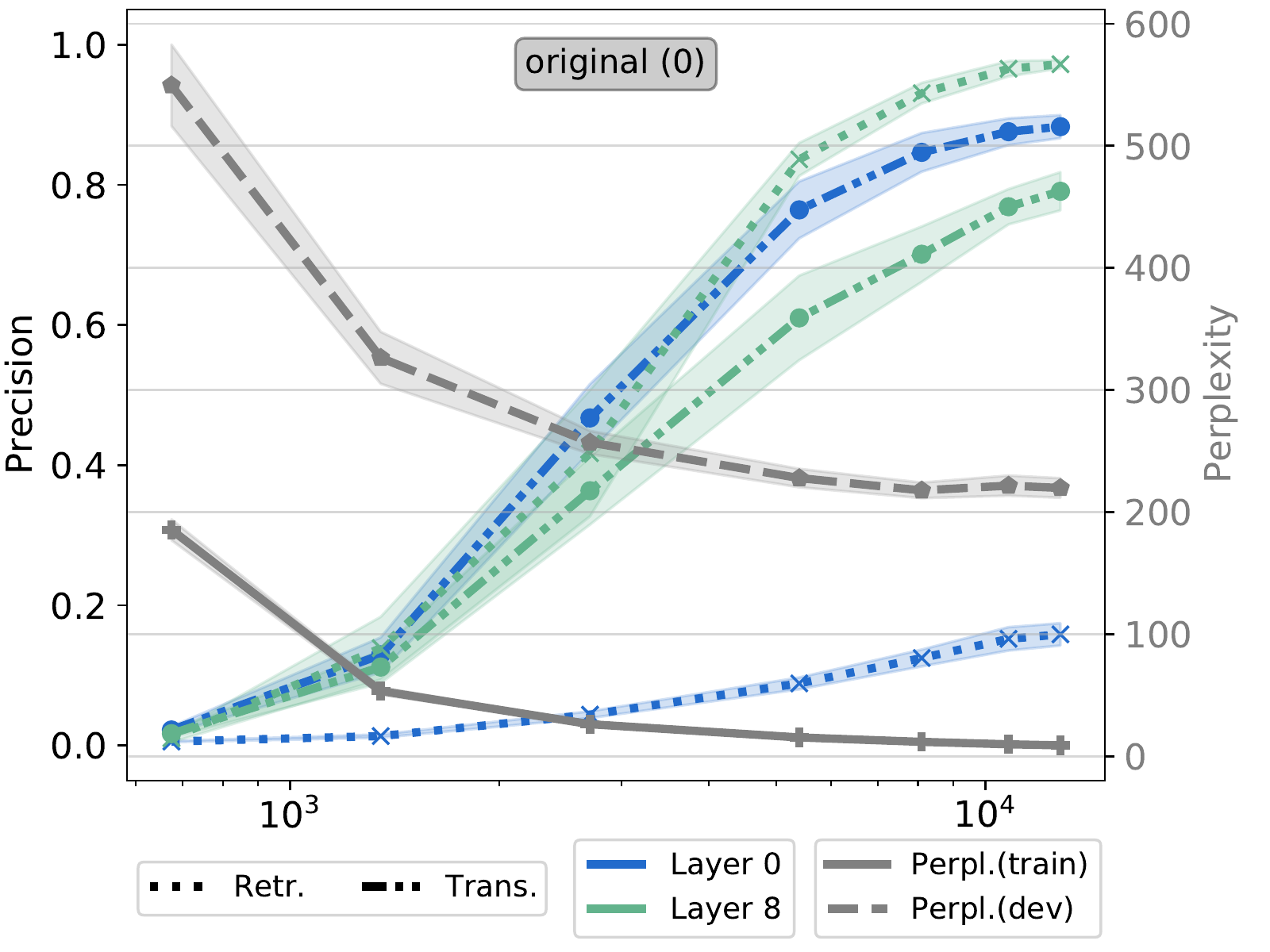}
	\includegraphics[width=0.45\linewidth]{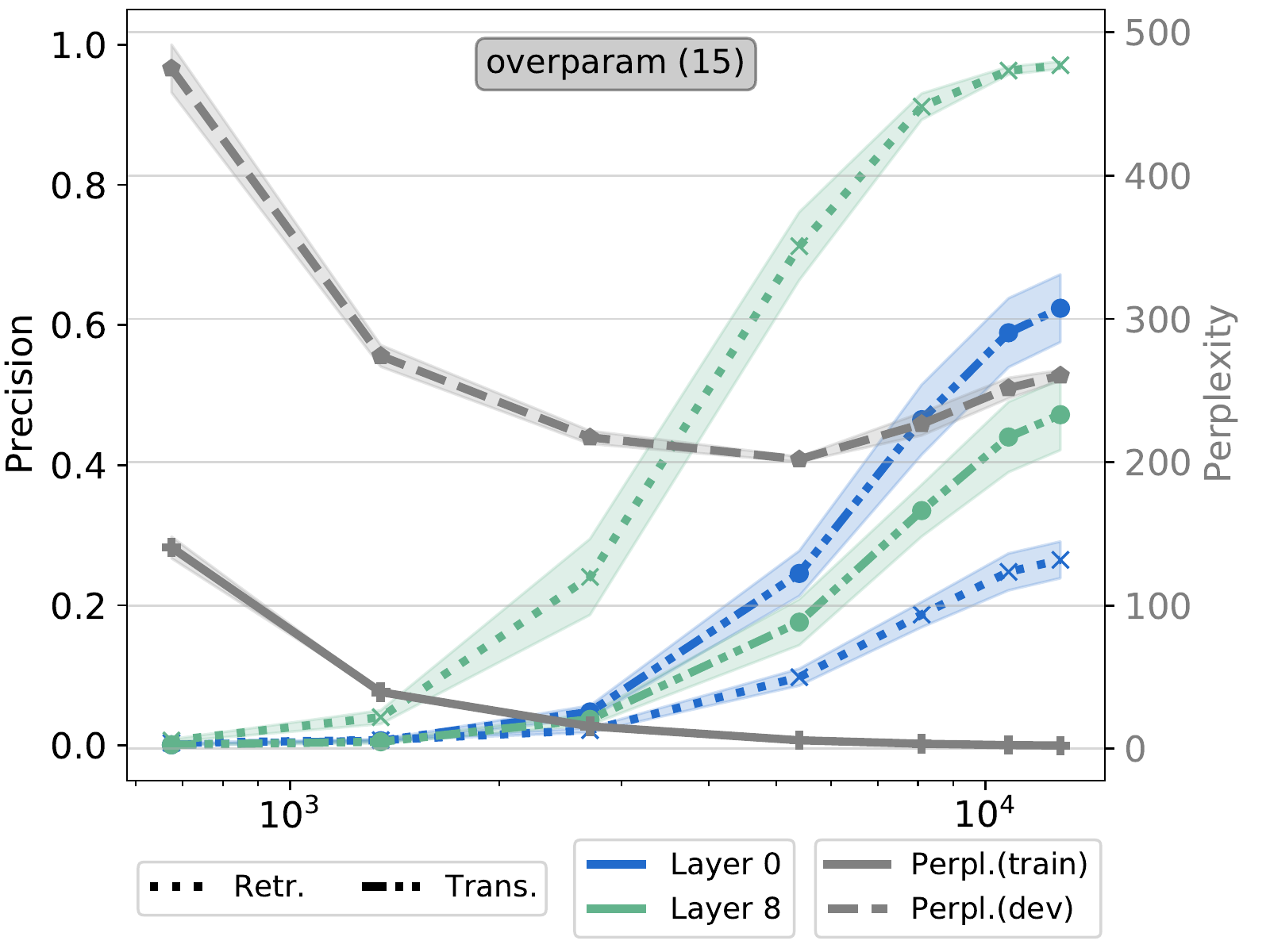}
	\caption{The longer a model is trained, the more multilingual it gets. x-axis shows
		training steps. Alignment $F_1$ is not shown as the models
		use shared position embeddings. Lines show mean and shaded areas show standard deviation across 5 random seed.}
	\figlabel{curve}
\end{figure*}

So far we have 
trained on a parallel corpus. Now we show
what happens with a merely comparable corpus.
The first half of the training corpus is used for English and the other half for Fake-English.
To mitigate the reduced amount of training data
we train for twice as many epochs. 
\tabref{parallel} shows that  multilinguality indeed decreases as the training 
corpus becomes non-parallel. This
suggests that the more comparable a training corpus is across
languages the higher the multilinguality. Note, however, 
that the models fit the training data worse and do not generalize as well as 
the original model.

\subsection{Multilinguality During Training}
\seclabel{overfit}

One central hypothesis is that BERT becomes multilingual at the
point at which it is forced to use its parameters 
efficiently. We argue that this point depends on several factors including the number of parameters, training duration, ``complexity'' of the data distribution and how easily common structures across language spaces can be aligned. The latter two are difficult to control for. We provided insights that two languages with identical structure but inverted word order are harder to align. \figref{curve} analyzes the former two factors and shows model fit and multilinguality for the small and large model settings over training steps. 

Generally, multilinguality rises very late at a stage where model fit improvements are flat. 
In fact, most of multilinguality in the overparameterized setting (15) arises once the model starts to overfit and perplexity
on the development set goes up. The original setting (0) has far fewer parameters. We hypothesize that it is forced to use its parameters efficiently and thus multilinguality scores rise much earlier when both training and development perplexity are still going down.

Although this is a very restricted experimental setup
it indicates that having multilingual models is a trade-off between 
good generalization and high degree of multilinguality. By
overfitting a model one could achieve high
multilinguality. \citet{conneau2019unsupervised} introduced
the concept of ``curse of multilinguality'' and found that the number
of parameters should be increased with the number of
languages.
Our results indicate that  too many parameters can also harm
multilinguality. However, in practice it is difficult to create a  model with so many parameters  that it is overparameterized when being trained on 104 Wikipedias.

\citet{ronnqvist2019multilingual} found that current multilingual BERT models may be undertrained. This is consistent with our findings that multilinguality arises late in the training stage.

\section{Improving Multilinguality}
 \seclabel{improving}

\begin{figure}[t]
	\centering
	\includegraphics[width=0.95\linewidth]{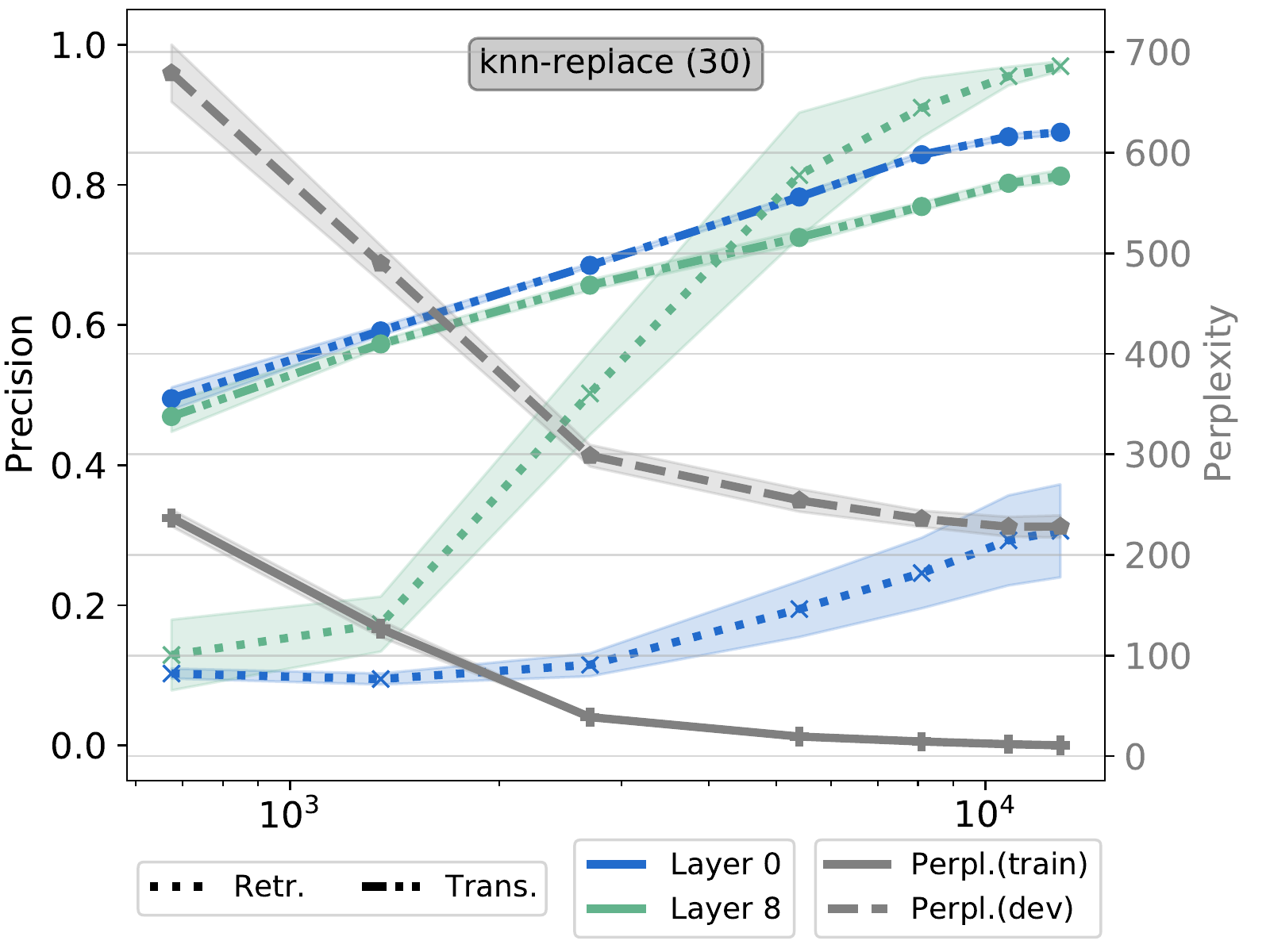}
	\caption{With knn-replace multilinguality rises earlier. Alignment $F_1$ is not shown as the model
		uses shared position embeddings.}
	\figlabel{curverepl}
\end{figure}

So far we have tried to break BERT's multilinguality. Now we
turn to exploiting our insights for improving it.
mBERT has shared position embeddings, shared special tokens
and we cannot change linguistic properties of languages.
Our results on overparameterization suggest that smaller models
become multilingual faster. However, 
mBERT 
may already be considered underparameterized given 
 that it is trained on 104 large Wikipedias. 
 
 One insight we can leverage
for the masking procedure
 is \emph{no-random}:  replacing masked words with random tokens. 
\emph{We propose to introduce a fourth masking option:} replacing masked tokens
 with semantically similar words from other languages. To this end we train
 static fastText embeddings \cite{bojanowski-etal-2017-enriching} on the training set and then project them into
 a common space using VecMap \cite{artetxe-etal-2018-robust}. We use this crosslingual space to replace masked
 tokens with nearest neighbors from the other language. Each masked word is then replaced with the probabilities 
 $(p_{\text{[mask]}}, p_{\text{[id]}}, p_{\text{[rand]}}, p_{\text{[knn]}}) = (0.5, 0.1, 0.1, 0.3)$, i.e., in $30\%$
 of the cases masked words get replaced with the nearest neighbor from the multilingual static embedding space.
 Note that this procedure (including VecMap) is fully unsupervised (i.e., no parallel data
 or dictionary required). We call this method \emph{knn-replace}. \newcite{wu2019emerging} performed similar experiments by creating code switched data and adding it to the training data. However, we only replace masked words.
 
 \figref{curverepl} shows the multilinguality score and model fit over training time. Compared to the original model in \figref{curve}, retrieval and translation have higher scores earlier.
 Towards the end multilinguality scores
 become similar, with knn-replace outperforming the original model (see \tabref{main}).
This finding is particularly important 
 for training BERT on large amounts of data. Given how
 expensive training is, it may not be possible to train a model
 long enough to obtain a high degree of
 multilinguality. Longer training incurs the risk of overfitting as well. 
 Thus
 achieving multilinguality early in the training process is
 valuable. Our new masking strategy has this property.

\section{Real Data Experiments}
\seclabel{real}

\subsection{XNLI}

We have presented experiments on a small corpus with English
and Fake-English.  Now we provide results on real data. Our
setup is similar to \cite{wang2019cross}: we train a
multilingual BERT model on English, German and Hindi. As
training corpora we sample 1GB of data from Wikipedia
(except for Hindi, as its size is $<$1GB ) and pretrain the
model for 2 epochs/140k steps with batch size 256 and learning rate 1e-4.  In this section,
we use BERT-base, not BERT-small because we found that
BERT-small with less than 1M parameters performs poorly in a
larger scale setup.  The remaining model and training
parameters are the same as before.  Each language has its
own vocabulary with size 20k.  We then evaluate the
pretrained models on XNLI \cite{conneau-etal-2018-xnli}. We
finetune the pretrained models on English XNLI (3 epochs, batch size 32, learning rate 2e-5,
following \citet{devlin-etal-2019-bert}).  Then the model is
evaluated on English. In addition, we do a zero-shot
evaluation on German and Hindi.

\tabref{xnli} presents accuracy on  XNLI test.
Compared to mBERT, accuracy is significantly lower but reasonable  on English ($.75$ vs. $.81$) -- 
we pretrain on far less data.
ID 0 shows high multilinguality with 0-shot accuracies $.57$ and $.45$. Inverting the order
of German has little effect on HIN, but DEU drops significantly (majority baseline is $.33$).
Our architectural modifications (8) harm both HIN and DEU. 
The proposed \emph{knn-replace}
model exhibits
the strongest degree of multilinguality, boosting the 0-shot
accuracy in DEU / HIN by 4\% / 9\%. Note that to accommodate  noise in the real world data, we randomly replace with one of the five nearest neighbors (not  the top nearest neighbor).
This indicates
that \emph{knn-replace} is useful for real world data and that
our prior findings transfer to larger scale settings.

\begin{table}[t]
	\centering
	\scriptsize
	\def\symmsep{0.1cm}
	\begin{tabular}{
			@{\hspace{\symmsep}}r@{\hspace{\symmsep}}
			@{\hspace{\symmsep}}l@{\hspace{\symmsep}}||
			@{\hspace{\symmsep}}r@{\hspace{\symmsep}}
			@{\hspace{\symmsep}}r@{\hspace{\symmsep}}
			@{\hspace{\symmsep}}r@{\hspace{\symmsep}}}
		\textbf{ID} & \textbf{Description} & \textbf{ENG} & \textbf{DEU} & \textbf{HIN} \\
		\midrule
0-base & original & \textbf{.75} \textsubscript{.00} & .57 \textsubscript{.02} & .45 \textsubscript{.01} \\
3-base & inv-order[DEU] &  \textbf{.75} \textsubscript{.00} & .41 \textsubscript{.01} & .46 \textsubscript{.04} \\
8-base & lang-pos;shift-special;no-random & .74 \textsubscript{.00} & .37 \textsubscript{.02} & .38 \textsubscript{.02} \\
30-base & knn-replace & .74 \textsubscript{.01} & \textbf{.61} \textsubscript{.01} & \textbf{.54} \textsubscript{.00} \\
\midrule
mBERT & Results by \cite{hu2020xtreme} & .81  & .70 & .59 \\
	\end{tabular}
	\caption{Accuracy on XNLI test for different model settings. Shown is the mean and standard deviation (subscript) across three random seeds. All models have the same architecture as BERT-base, are pretrained on Wikipedia data and
		finetuned on English XNLI training
                data. mBERT was pretrained longer and on
                much more data and has thus higher
                performance. Best non-mBERT performance in bold.\tablabel{xnli}}
\end{table}

\section{Related Work}

There is a range of prior work analyzing the reason for BERT's multilinguality.
\citet{singh2019bert} show that 
BERT stores language representations in different subspaces and investigate how subword tokenization influences multilinguality. 
\citet{artetxe2019cross} show that neither a shared vocabulary nor joint pretraining is 
essential for multilinguality.
\citet{wang2019cross} extensively study reasons for multilinguality (e.g., researching depth,
number of parameters and attention heads). They conclude that 
depth is essential. 
They also investigate
language properties and conclude that 
structural similarity across languages is important, without further defining this term. 
Last, \citet{wu2019emerging} find that a shared vocabulary is not required. They find that 
shared parameters in the top layers are required for multilinguality. Further they show that 
different monolingual BERT models exhibit a similar structure and thus conclude that mBERT somehow
aligns those isomorphic spaces. They investigate having separate embedding look-ups per language (including position embeddings and special tokens) and a variant of avoiding
cross-language replacements. Their method ``extra anchors'' yields a higher degree of multilinguality.
In contrast 
to this prior work, we investigate multilinguality in a clean laboratory setting, 
investigate the interaction of architectural aspects and 
research new aspects such as overparameterization or inv-order. 

Other work focuses on creating better multilingual models. 
\citet{mulcaire-etal-2019-polyglot} proposed a method to learn multilingual contextual representations.
\citet{conneau2019cross} introduce the translation modeling objective. 
\citet{conneau2019unsupervised} propose XLM-R. They introduce the term 
``curse of multilinguality'' and show that multilingual model quality degrades with an increased number of 
languages given a fixed number of parameters. This can be interpreted as the minimum number of parameters required whereas we find indications
that models that are too large can be harmful for multilinguality as well.
\citet{cao2020multilingual} improve the multilinguality of mBERT by introducing a regularization term 
in the objective, similar to the creation of static multilingual embedding spaces. 
\citet{huang2019unicoder} extend mBERT pretraining with three additional tasks and show an improved overall performance. More recently, better multilinguality is achieved by \citet{pfeiffer2020mad} (adapters) and \citet{chi2020infoxlm} (parallel data).
We propose a simple extension to make mBERT more
multilingual; it does not require additional supervision, parallel data
or a more complex loss function -- in contrast to this prior work.

Finally, many papers
find that mBERT yields competitive zero-shot performance across a  range of languages 
and tasks such as parsing and NER
\cite{pires-etal-2019-multilingual,wu2019beto}, word alignment and sentence retrieval \cite{libovicky2019language} and
language generation
\cite{ronnqvist2019multilingual};
\citet{hu2020xtreme} show this for
40 languages and 9 tasks. \citet{wu-dredze-2020-languages}
consider the performance on up to 99 languages for NER. In contrast, \citet{lauscher2020zero} show limitations of the zero-shot setting and \citet{zhao-etal-2020-limitations} observe poor performance of mBERT in reference-free machine translation evaluation.
Prior work here focuses on investigating the degree of multilinguality, not the reasons for it.

\section{Conclusion}

We investigated which architectural and linguistic properties 
are essential for BERT to yield 
crosslingual representations. The main takeaways are: 
\textbf{i)} Shared position embeddings, shared special tokens, replacing masked
tokens with random tokens and a limited amount of parameters are necessary elements
for multilinguality.
\textbf{ii)} Word order is relevant: BERT is not multilingual with one language having an 
inverted word order.
\textbf{iii)} The comparability of training corpora contributes to multilinguality.  
We show that our findings transfer to larger scale settings.
We experimented with a simple
modification to obtain stronger multilinguality in BERT
models and demonstrate its effectiveness on XNLI. We considered a fully unsupervised setting without any crosslingual signals. In future work we plan to incorporate crosslingual signals as \citet{vulic-etal-2019-really} argue that a fully unsupervised setting is hard to motivate.

\section*{Acknowledgements}
We gratefully
acknowledge
funding through a
Zentrum Digitalisierung.Bayern 
fellowship awarded to the first author. This work was supported
by the European Research Council (\# 740516). 
We thank Mengjie Zhao, Nina P\"orner, Denis Peskov and the anonymous reviewers for fruitful discussions and valuable comments.

\bibliography{anthology,multl}
\bibliographystyle{acl_natbib}

\appendix

\section{Additional Details on Methods}
\seclabel{methods}

\subsection{Word Translation Evaluation}

Word translation is evaluated in the same way as sentence retrieval. This section provides additional details.

For each token in the vocabulary $w^{(k)}$ we feed the ``sentence'' ``[CLS] \{ $w^{(k)}$\} [SEP]'' to the BERT model 
to obtain the embeddings $\mathcal{E}(w^{(k)}) \in \mathbb{R}^{3 \times d}$ from the $l$-th layer of BERT
for $k \in \{\text{eng}, \text{fake}\}$. Now, we extract the word embedding by taking
the second vector (the one corresponding to  $w^{(k)}$) and denote it by 
$e_w^{(k)}$. Computing cosine 
similarities between English and Fake-English tokens
yields the similarity matrix $R \in \mathbb{R}^{m \times m}$ 
where $R_{ij} = \text{cosine-sim}(e_i^{(\text{eng})}, e_j^{(\text{fake})})$
for $m$ tokens in the vocabulary of one language (in our case $2048$).

Given an English query token $s_i^{(\text{eng})}$, we obtain the retrieved tokens in Fake-English by 
ranking them according to similarity. Note that we can do the same with Fake-English
as query language. We report the mean precision of these directions that is computed as
$$
\tau = \frac{1}{2m}\sum_{i = 1}^{m}\mathds{1}_{\arg\max_{l}R_{il} = i} + \mathds{1}_{\arg\max_{l}R_{li} = i}.
$$

\subsection{inv-order}

Assume the sentence ``He ate wild honey .'' exists in the corpus. 
The tokenized version is [He, ate, wild, hon, \#\#e, \#\#y, .] and the corresponding Fake-English sentence 
is [::He, ::ate, ::wild, ::hon, ::\#\#e, ::\#\#y, ::.]. If we apply the modification inv-order we always invert the order of the 
Fake-English sentences, thus the model only receives the sentence  [::., ::\#\#y, ::\#\#e, ::hon, ::wild, ::ate, ::He].

\subsection{knn-replace}
We use the training data to train static word embeddings for each language using the tool 
fastText. Subsequently we use VecMap \cite{artetxe-etal-2018-robust} to map the embedding spaces from each language into the English 
embedding space, thus creating a multilingual static embedding space. We use VecMap without any supervision. 

During MLM-pretraining of our BERT model 15\% of the tokens are randomly selected and ``masked''. They then get either replaced by 
``[MASK]'' (50\% of the cases), remain the same (10\% of the cases), get replaced by a random other token (10\% of the cases) or we replace 
the token with one of the five nearest neighbors (in the fake-English setup only with the nearest neighbor) from another language (30\% of the cases). Among those five nearest neighbors we pick one randomly. In case more than one other language is available we pick one randomly.

\section{Additional Non-central Results}
\seclabel{results}

\subsection{Model 17}

One might argue that our model 17 in Table 1 of the main paper is simply not trained enough
and thus not multilingual.
However, \tabref{overfit} shows that even when continuing to train this model for a long time
no multilinguality arises. Thus in this configuration the model has enough capacity to model the languages independently of each other --
and due to the modifications apparently no incentive to try to align the language representations.

\subsection{Word Order in XNLI}

\begin{table}[t]
	\centering
	\scriptsize
	\begin{tabular}{r|rr}
		Lang. & Kendall's Tau Distance & XNLI Acc.\\
		\midrule\midrule
		en &1.0 &  81.4\\ \midrule
		ar & 0.72 & 64.9\\
		de & 0.74 & 71.1\\
		fr & 0.80 & 73.8\\
		ru & 0.72 & 69.0\\
		th & 0.71 & 55.8\\
		ur & 0.59 & 58.0\\
		zh & 0.68 & 69.3\\
		bg & 0.75 & 68.9\\
		el & 0.77 & 66.4\\
		es & 0.76 & 74.3\\
		hi & 0.58 & 60.0\\
		sw & 0.73 & 50.4\\
		tr & 0.47 & 61.6\\
		vi & 0.78 & 69.5\\
	\end{tabular}
	\caption{Kendall's Tau word order metric and XNLI zero-shot accuracies. \tablabel{synchronity}}
\end{table}

To verify whether similar word order across languages influences the multilinguality
we propose to compute a word reordering metric and correlate this metric with the performance
of 0-shot transfer capabilities of mBERT. To this end we consider the performance
of mBERT on XNLI. 
We follow \citet{birch-osborne-2011-reordering} in computing word reordering metrics
between parallel sentences (XNLI is a parallel corpus). More specifically we compute the 
Kendall's tau metric. To this end, we compute word alignments 
between two sentences using the Match algorithm by \citet{sabet2020simalign}, which directly
yield a permutation between sentences as required by the distance metric. 
We compute the metric on 2500 sentences from the development data of XNLI and average it across sentences
to get a single score per language. The scores and XNLI accuracies are in \tabref{synchronity}.

The Pearson correlation between Kendall's tau metric and the XNLI classification accuracy in a zero-shot 
scenario (mBERT only finetuned on English and tested on all other languages) is 46\% when disregarding English and 64\% when including English.  Thus there is a some correlation observable. This indicates that zero-shot performance of mBERT might also rely on similar word order across languages. 
We plan to extend this experiment 
to more zero-shot results and examine this effect more closely in future work.

\subsection{Larger Position Similarity Plots}
We provide larger versions of our position similarity plots in \figref{posfig}.

\section{Reproducibility Information}
\seclabel{repro}

\subsection{Data}

\tabref{data} provides download links to data.

\subsection{Technical Details}

\begin{table}[t]
	\scriptsize
	\centering
	\begin{tabular}{p{5cm}|r}
		Scenario & Runtime \\
		\midrule
		pretrain small BERT model on Easy-to-Read-Bible, 100 epochs & $\sim35m$ \\
		\midrule
		pretrain large BERT model (BERT-base) on Easy-to-Read-Bible, 100 epochs  & $\sim4h$ \\\midrule
		pretrain large BERT model (BERT-base) on Wikipedia sample, 1 epoch  & $\sim2.5 days$  \\
	\end{tabular}
	\caption{Runtime on a single GPU.\tablabel{runtimes}}
\end{table}

\begin{table}[t]
	\scriptsize
	\centering
	\begin{tabular}{l|r}
		Model & Parameters \\
		\midrule
		Standard Configuration (``Small model'') & $1M$ \\
		BERT-Base / Overparameterized Model / ``Large model'' & $88M$ \\
		Real data model (BERT-Base with larger vocabulary) & $131M$ \\
		mBERT & $178M$
	\end{tabular}
	\caption{Number of parameters for our used models.\tablabel{nparams}}
\end{table}

The number of parameters for each model are in \tabref{nparams}.

We did all computations on a server with up to 40 Intel(R) Xeon(R) CPU E5-2630 v4 CPUs
and 8 GeForce GTX 1080Ti GPU with 11GB memory. No multi-GPU training was performed. 
Typical runtimes are reported in \tabref{runtimes}.

Used third party systems are shown in \tabref{party}.

\subsection{Hyperparameters}

We show an overview on hyperparameters in \tabref{hyperparams}. If not shown we fall back to default values in the systems.

\begin{table*}[h]
	\centering
	\scriptsize
	\centering
	\begin{tabular}{p{2cm}p{1cm}p{3cm}p{3cm}p{5cm}}
		Name & Languages & Description & Size & Link \\
		\midrule
		\midrule
		XNLI \cite{conneau-etal-2018-xnli} & English, German, Hindi & Natural Language Inference Dataset. We use the English training set and
		English, German and Hindi test set. & 392703 sentence pairs in train, 5000 in test, 2500 in dev per language. &  \url{https://cims.nyu.edu/~sbowman/xnli/} \\
		\midrule
		Wikipedia &English, German, Hindi& We use 1GB of randomly sampled data from a Wikipedia dump downloaded in October 2019. &8.5M sentences for ENG, 9.3M for DEU and 800K for HIN.& \url{download.wikimedia.org/[X]wiki/latest/[X]wiki-latest-pages-articles.xml.bz2}  \\
		\midrule
		Bible \cite{mayer2014creating} &English& We use the editions Easy-to-Read and King-James-Version. &We use all 17178 sentences in Easy-to-Read (New Testament) and the first 10000 sentences of King-James in the Old Testament. & n/a\\
	\end{tabular}
	\caption{Overview on datasets.\tablabel{data}}
\end{table*}

\begin{table*}[h]
	\scriptsize
	\centering
	\begin{tabular}{c||lp{10cm}}
		\textbf{System} & \textbf{Parameter} & \textbf{Value} \\
		\midrule
		\multirow{3}{50pt}{Vecmap}
		&Code URL& \url{https://github.com/artetxem/vecmap.git}\\
		&Git Commit Hash& b82246f6c249633039f67fa6156e51d852bd73a3\\
		&&\\
		\multirow{4}{50pt}{fastText}
		&Version &0.9.1 \\
		&Code URL & \url{https://github.com/facebookresearch/fastText/archive/v0.9.1.zip}\\
		&Embedding Dimension&300\\
		\multirow{1}{50pt}{Transformers} & Version& 2.8.0\\
		\multirow{1}{50pt}{Tokenizers}  &Version &0.5.2\\
		\multirow{1}{50pt}{NLTK}  &Version&3.4.5\\
	\end{tabular}
	\caption{Overview on third party systems used. }
	\tablabel{party}
\end{table*}

\begin{table*}[b]
	\scriptsize
	\centering
	\begin{tabular}{lp{10cm}}
		\textbf{Parameter} & \textbf{Value}\\
		\midrule
		Hidden size & $64$; $768$ for large models (i.e., overparameterized and those used for XNLI) derived from BERT-based configuration\\
		Intermediate layer size & $256$; $3072$ for large models \\
		Number of attention heads & $1$; $12$ for large models \\
		Learning rate & $2e-3$ (chosen out of $1e-4$, $2e-4$, $1e-3$, $2e-r$, $1e-2$, $2e-2$ via grid search; criterion: perplexity); $1e-4$ for large models, same as used in \cite{devlin-etal-2019-bert}\\
		Weight decay & $0.01$ following \cite{devlin-etal-2019-bert} \\
		Adam epsilon & $1e-6$ following \cite{devlin-etal-2019-bert}  \\
		Random Seeds & 0, 42, 43, 100, 101; For single runs: 42. For real data experiments: 1,42 and 100.\\
		Maximum input length after tokenization & 128 \\
		Number of epochs & 100 unless indicated otherwise.  (chosen out of $10$, $20$, $50$, $100$, $200$ via grid search; criterion: perplexity)\\
		Number of warmup steps & 50 \\
		Vocabulary size & 4096; 20000 per language for the XNLI models \\
		Batch size & 256 for pretraining (for BERT-Base models 16 with 16 gradient accumulation steps), 32 for finetuning\\
	\end{tabular}
	\caption{Model and training parameters during pretraining.}
	\tablabel{hyperparams}
\end{table*}

\begin{table*}[b]
	\centering
	\scriptsize
	\def\symmsep{0.15cm}
	\centering
	\begin{tabular}{
			@{\hspace{\symmsep}}r@{\hspace{\symmsep}}
			@{\hspace{\symmsep}}l@{\hspace{\symmsep}}
			@{\hspace{\symmsep}}r@{\hspace{\symmsep}}||
			@{\hspace{\symmsep}}r@{\hspace{\symmsep}}|
			@{\hspace{\symmsep}}r@{\hspace{\symmsep}}
			@{\hspace{\symmsep}}r@{\hspace{\symmsep}}
			@{\hspace{\symmsep}}r@{\hspace{\symmsep}}|
			@{\hspace{\symmsep}}r@{\hspace{\symmsep}}
			@{\hspace{\symmsep}}r@{\hspace{\symmsep}}
			@{\hspace{\symmsep}}r@{\hspace{\symmsep}}|
			@{\hspace{\symmsep}}r@{\hspace{\symmsep}}
			@{\hspace{\symmsep}}r@{\hspace{\symmsep}}}
		&&&  \textbf{Mult.-} & \multicolumn{3}{c}{ \textbf{Layer 0} } & \multicolumn{3}{c}{ \textbf{Layer 8} }  & \multicolumn{2}{c}{ \textbf{MLM-} } \\
		&& \textbf{Num.}& \textbf{score} & \textbf{Align.} & \textbf{Retr.} & \textbf{Trans.} & \textbf{Align.} & \textbf{Retr.} & \textbf{Trans.} &\multicolumn{2}{c}{ \textbf{Perpl.} }  \\
		\textbf{ID}&\textbf{Description} & \textbf{Epochs}& \textbf{$\mu$} & \textbf{$F_1$} & \textbf{$\rho$} & \textbf{$\tau$} &  \textbf{$F_1$} & \textbf{$\rho$} & \textbf{$\tau$} &\textbf{train} & \textbf{dev}\\
		\midrule
		0 & original & 100 & .70 & 1.00 \textsubscript{.00} & .16 \textsubscript{.02} & .88 \textsubscript{.02} & 1.00 \textsubscript{.00} & .97 \textsubscript{.01} & .79 \textsubscript{.03} & 9 \textsubscript{00.22} & 217 \textsubscript{07.8} \\
		17 & lang-pos;shift-special;no-random;overparam & 100 & .00 & .05 \textsubscript{.02} & .00 \textsubscript{.00} & .00 \textsubscript{.00} & .05 \textsubscript{.04} & .00 \textsubscript{.00} & .00 \textsubscript{.00} & 2 \textsubscript{00.02} & 270 \textsubscript{20.1}  \\
		17 & lang-pos;shift-special;no-random;overparam & 250 & .00 & .06 \textsubscript{.02} & .00 \textsubscript{.00} & .00 \textsubscript{.00} & .06 \textsubscript{.05} & .00 \textsubscript{.00} & .00 \textsubscript{.00} & 1 \textsubscript{00.00} & 1111 \textsubscript{30.7} \\
	\end{tabular}
	\caption{Even when continuing the training for a long time 
		overparameterized models with architectural
		modifications do not become
		multilingual.\tablabel{overfit}}
\end{table*}

\begin{figure*}[h]
	\centering
	\includegraphics[width=0.7\linewidth]{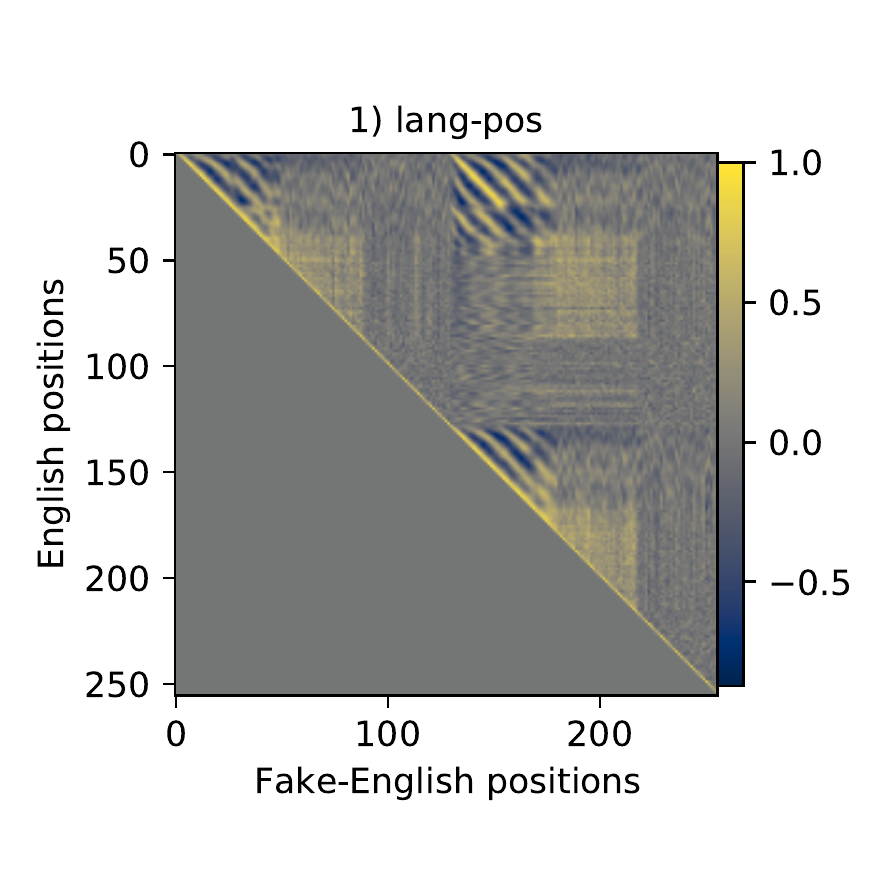}
	\includegraphics[width=0.7\linewidth]{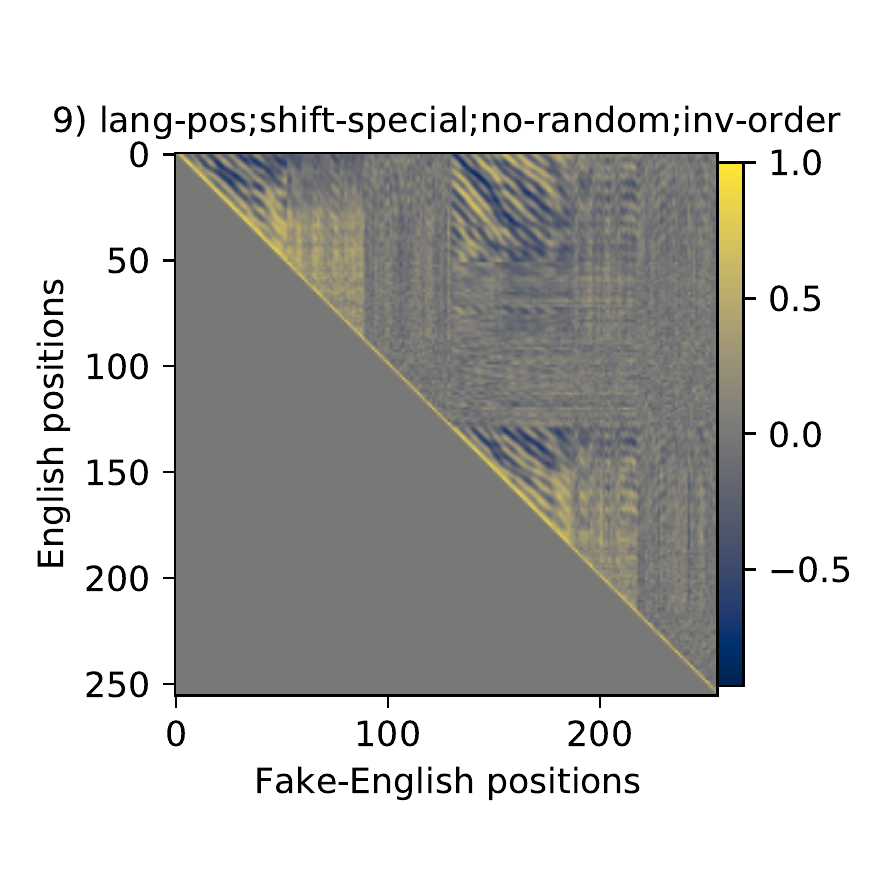}
	\caption{Cosine similarity of position embeddings. IDs 0-127 are used for English, 128-255 for Fake-English. \figlabel{posfig}.}
\end{figure*}

\end{document}